\documentclass[11pt]{article}

\usepackage{hyperref}
\usepackage{latexsym}
\usepackage{graphicx}
\usepackage{epsfig}
\usepackage{xurl}
\usepackage{amsmath,amssymb}
\numberwithin{equation}{section}
\usepackage{tabularx}
\usepackage{authblk}
\usepackage{mathrsfs}
\usepackage{algorithm}
\usepackage{soul}
\usepackage{multirow}
\usepackage{color}
\usepackage{subcaption}
\usepackage{enumerate}
\usepackage{booktabs}
\usepackage{svg}
\usepackage{cleveref}
\usepackage{xcolor}
\usepackage{capt-of}
\usepackage{makecell}
\usepackage{cite}
\usepackage[table]{xcolor}
\usepackage[noend]{algpseudocode}
\usepackage[utf8]{inputenc}

\newcommand{\name}{ESGPred}

\title{A Modular Framework for ESG Prediction through Domain-Specific Language Modeling and Graph Learning}

\author{
Sofia Baiocco$^{1}$,
Gianluca Bonifazi$^{2,*}$,
Christopher Buratti$^{2}$,
Michele Marchetti$^{2}$,
Camilla Mazzoli$^{1}$,
Federica Parlapiano$^{2}$,
Giulia Quaglieri$^{2}$,
Davide Traini$^{2,3}$,
Domenico Ursino$^{2}$,
Luca Virgili$^{2}$ \\

$^{1}$ DIMA, Polytechnic University of Marche, Italy \\
$^{2}$ DII, Polytechnic University of Marche, Italy \\
$^{3}$ CHIMOMO, University of Modena and Reggio Emilia, Italy \\
$^{*}$ Corresponding author \\

s.baiocco@pm.univpm.it;
g.bonifazi@univpm.it;
c.buratti@pm.univpm.it;
michele.marchetti@univpm.it;
c.mazzoli@univpm.it;
f.parlapiano@pm.univpm.it;
g.quaglieri@pm.univpm.it;
davide.traini@unimore.it;
d.ursino@univpm.it;
luca.virgili@univpm.it
}

\date{}

\begin{document}
\maketitle

\begin{abstract}
Environmental, Social and Governance (ESG) scores are increasingly used to evaluate corporate sustainability, supporting investment decisions, ensuring regulatory compliance and assessing risk. However, current ESG evaluations largely rely on proprietary methodologies that often produce inconsistent ratings. This motivates the development of data-driven prediction frameworks. In this paper, we present \name{}, a modular framework for predicting ESG scores by combining domain-specific language modeling with graph-based learning over a network that models the relationships among the companies being evaluated. In addition to presenting the rationale behind \name{} and detailing its ESG score prediction process, we show that \name{} can effectively predict both overall ESG ratings and their individual Environmental, Social, and Governance components. We evaluated \name{} using a dataset of 5,266 sustainability reports published by 2,134 publicly listed companies between 2020 and 2023. In this evaluation, we used ESG ratings provided by LSEG Data \& Analytics as ground truth. Our experimental results demonstrated that integrating graph-based relational learning improves prediction performance over purely text-based baselines. 
\end{abstract}


\section{Introduction}
\label{sec:Introduction}

In recent years, Environmental, Social and Governance (ESG) criteria have become central to sustainable finance, corporate governance, and institutional investment decisions \cite{YuCaZh26, Vo*19}. ESG scores summarize a company's sustainability performance beyond traditional financial indicators and influence capital allocation, risk assessment, and stakeholder evaluation \cite{RaYu24, Liu*26, MaKrFo24}. Meanwhile, regulatory attention has intensified, particularly in the European Union, where the Corporate Sustainability Reporting Directive (CSRD)\footnote{\url{https://finance.ec.europa.eu/capital-markets-union-and-financial-markets/company-reporting-and-auditing/company-reporting/corporate-sustainability-reporting_en}} aims to improve the transparency, comparability, and standardization of corporate sustainability reports. Similar initiatives in other jurisdictions reflect a broader global shift toward more accountable, data-driven ESG reporting practices \cite{Fagbemi*25}.

Despite this growing institutional and regulatory relevance, ESG assessment remains challenging and only partially standardized \cite{ShYa25}. ESG ratings are commonly produced by specialized companies, such as Sustainalytics\footnote{\url{https://www.sustainalytics.com/}}, Moody's Vigeo-Eiris\footnote{\url{https://www.moodys.com/}}, S\&P Global\footnote{\url{https://www.spglobal.com/}}, MSCI\footnote{\url{https://www.msci.com/}}, and LSEG Data \& Analytics\footnote{\url{https://www.lseg.com/}}. These agencies rely on proprietary methodologies that differ in terms of data sources, rating scales, weighting schemes, and aggregation criteria. Consequently, the same company may receive substantially different ESG evaluations depending on the agency \cite{BeKoRi22}. This lack of alignment raises concerns about the comparability, reliability, and transparency of ESG scores, particularly among investors, regulators, and stakeholders who use them for decision-making purposes.

The main reason for this discrepancy is the challenge of translating diverse, mainly qualitative corporate reports into consistent, quantitative indicators. Although sustainability reports contain valuable information about environmental policies, social initiatives, governance structures, risk management practices, and compliance strategies, they are lengthy and written in specialized corporate and financial language. Manual assessment is consequently costly and difficult to reproduce. 

Conversely, fully automated approaches can extract meaningful patterns from large volumes of unstructured textual data. Recent advances in Natural Language Processing (NLP) have demonstrated the effectiveness of Transformer-based models in analyzing financial and corporate documents. FinBERT \cite{Liu*21-7}, a BERT-based model pre-trained on financial corpora, is particularly suitable for financial text-mining tasks because it provides language representations well-suited to them. However, ESG prediction cannot be treated as a purely textual problem. In fact, companies often share sector-specific characteristics, business models, regulatory constraints, organizational structures, markets, products, and sustainability narratives. Ignoring these relationships may limit the predictive models' ability to capture broader patterns of ESG behavior. Graph-based learning can provide a great support to handle these relational dependencies. In particular, it is possible to define a graph whose nodes represent companies and/or their sustainability reports and whose edges denote the corresponding relationships. We refer to this structure as a ``sustainability graph'' in the following. Given one or more sustainability graphs, Graph Neural Networks (GNNs) can be employed to detect information and knowledge about the ESG sustainability of the corresponding companies, for instance to determine ESG scores or classify companies into suitable ESG classes. In the ESG domain, this perspective is particularly relevant because relationships between companies can be multiple and heterogeneous. For instance, they can involve the companies' industrial sector, the semantic content of their sustainability reports, their business activity, and so forth. 

In this paper, we present \name{}, a modular framework that combines domain-specific language modeling with graph-based learning to predict the ESG class of a company. This class is encoded as a categorical variable corresponding to ESG score quartiles. To perform its task, first \name{} fine-tunes FinBERT in a supervised setting using ESG-related labels in order to adapt the FinBERT's semantic representations to the sustainability domain. Since sustainability reports often exceed the token limit of Transformer architectures, documents are segmented into chunks, which are then embedded and aggregated to obtain a representation for each sustainability report. \name{} uses these embeddings to construct some sustainability graphs. In particular, it constructs one sustainability graph for each kind of relationship of interest; for instance, it can construct a sustainability graph considering the semantic similarity of sustainability reports, another capturing the similarity of company descriptions, a third one taking the company's industrial sector into account, and so on. Finally, \name{} leverages the resulting sustainability graphs to train a Graph Neural Network model, which combines the textual information of each report with knowledge derived from sustainability graphs for determining the ESG class of a company.

 Unlike approaches based solely on structured financial indicators or document analysis, \name{} allows for the systematic evaluation of the impact of sustainability reports. This is achieved by using several sustainability graphs, each considering a specific kind of relationship between the involved sustainability reports and/or companies, as well as graph learning models and ESG targets. Experimental results on a large-scale dataset of 5,266 sustainability reports published by 2,134 publicly listed companies from 2020 to 2023 demonstrate the benefits of combining both language modeling and graph learning. 

The rest of this paper is organized as follows. Section \ref{sec:Related-Work} reviews the related literature on ESG prediction, NLP-based sustainability assessment, and graph-based methods. Section \ref{sec:Proposed-Approach} provides a technical description of \name{}. Section \ref{sec:Experiments} describes the experiments performed, the obtained results and the hyperparameter tuning. Section~\ref{sec:Discussion}  discusses the strengths, limitations, and implications of \name{}. Finally, Section \ref{sec:Conclusion} summarizes the main findings and outlines some future research directions.

\section{Related Work}
\label{sec:Related-Work}

ESG scores are widely recognized in current literature as a standard metric for corporate sustainability that extends beyond conventional financial indicators \cite{JoMa23}, \cite{Halid*23}, \cite{Cohen23}. Private rating agencies calculate these scores using proprietary methods that combine human expertise and automated data processing \cite{EsMuFe10}. To mitigate inconsistent and subjective outcomes, research is shifting toward data-driven, automated ESG evaluation frameworks. A comprehensive review underscores this shift from qualitative analyst assessments to advanced, data-driven frameworks \cite{ZhZh24}. In this context, Artificial Intelligence (AI) in general, and Machine Learning (ML) in particular, are essential for processing diverse datasets and integrating structured financial records and unstructured textual data.

A significant portion of recent ESG literature attempts to predict sustainability scores using structured financial data \cite{Taskin*25}, \cite{Sariyer*24}. These studies exploit the relationship between a firm's economic indicators and its ESG performance \cite{DaDeLe21}, \cite{Martin-Melero*25}. To capture these complex dynamics, researchers are replacing traditional linear regression with robust nonlinear methodologies \cite{Chowdhury*23}, \cite{DaDeLe22}. For example, the author of \cite{Chaouech25} proposes a framework based on advanced gradient boosting algorithms, such as CatBoost, to reveal intricate connections between financial and environmental indicators. As a further advance in this direction, the authors of \cite{KrBoBo21} present an ensemble model combining neural networks with XGBoost and CatBoost. Their research shows that basic financial metrics can reliably estimate initial ESG scores, which is beneficial for small companies that lack comprehensive sustainability reporting. 

Further studies shift the focus to geographical contexts and investigate regional market variations \cite{ZhZh26}, \cite{ChChLe24}, \cite{AlYi26}. For instance, the authors of  \cite{LiHs23} demonstrate that governance and financial data are highly effective ESG predictors for Taiwanese non-financial companies. They emphasize that Asian markets present distinct liquidity risks and a higher density of technology companies than Western economies. Similarly, the authors of \cite{David*24} design a model customized for the Chinese market. They demonstrate that Machine Learning approaches require adaptation to local regulatory contexts to accurately reflect regional sustainability dynamics. 

Despite recent methodological progress, the authors of \cite{PaNaDe26} point out that financial metrics alone cannot fully capture ESG performance. To address this limitation, they propose a stacked ensemble model that integrates structured non-financial variables, such as controversy and geospatial indicators.

However, relying solely on structured data is inherently limiting. A comprehensive understanding of corporate responsibility requires analyzing unstructured data \cite{Pawar*25}, \cite{Fischbach*23}, \cite{Alarfaj26}. Consequently, a substantial portion of recent research has shifted toward Natural Language Processing (NLP) \cite{Ngee*24}, \cite{Schimanski*24}. These studies posit that corporate narratives, such as Corporate Social Responsibility (CSR) statements and annual reports, contain crucial information for predicting ESG performance, which traditional structured financial data fail to capture \cite{Sokolov*21}, \cite{LeKi23}. 

A significant obstacle to the linguistic analysis of company sustainability reports is that general-purpose language models cannot accurately process specialized financial terminology. To address this issue, the authors of \cite{LeKiJu25} propose ESG-KIBERT, a domain-specific variant of BERT that leverages a hard attention mechanism to improve semantic interpretation. Integrating this architecture with FinBERT effectively allows for the evaluation of the thematic content of corporate ESG narratives alongside broader market sentiment, capturing the interplay between official reporting and investor perception. 

Furthermore, company reports are often structured to minimize negative sentiment, so standard NLP tools have difficulty identifying underlying sustainability risks. To address this issue, the authors of \cite{PaEh22} use Transformer-based models trained on official company filings. These models relate textual representations to established ESG ratings within a binary classification framework. The results described in \cite{PaEh22} suggest that pre-screening ESG-relevant passages allows these models to significantly outperform traditional methods when classifying corporate sustainability performance. The authors of \cite{Zhang*25-2} introduce E-BERT, which automates the evaluation pipeline by fine-tuning BERT on a specialized corpus. E-BERT filters out non-essential information in corporate documents, enabling the model to distinguish authentic sustainability initiatives from promotional rhetoric. This specialized focus allows E-BERT to consistently outperform baseline deep learning methods. The ESG2PreEM framework, proposed in \cite{Lee*24-1}, evaluates textual data by integrating FinBERT sentiment features with a soft-voting ensemble architecture that includes BERT, ALBERT, and RoBERTa. 

Graph-based methods for modeling and exploiting company relationships have gained traction in ESG research. The authors of \cite{Bernhardt*25} model company governance as a network connecting companies through shared decision-makers. They then use GraphSAGE \cite{HaYiLe17} and GAT \cite{Velickovic*18} architectures to integrate node-level features and network structure, achieving stronger classification results. Similarly, the authors of \cite{AnGuLi23} predict ESG ratings by constructing dynamic graphs from multiple sources, such as overlapping board members, mutual fund co-ownership, and media co-mentions. This enriches the model with centrality-based features.

Despite these advances, applying Graph Neural Networks (GNNs) to predict ESG scores is still in its early stages. The existing literature largely focuses on explicit, observable inter-company relationships, such as operational or strategic linkages. The \name{} framework introduced in this paper systematically examines multiple forms of report and/or company relationships and integrates them with text-based analysis of sustainability reports to obtain more complete and refined ESG scores. Furthermore, \name{} enables the automated assessment of ESG performance based on corporate disclosures and context, thus addressing the consistency issues that could arise from a human-based evaluation performed by agencies, each having a proprietary procedure.

\section{Proposed Approach}
\label{sec:Proposed-Approach}

\name{} uses textual information associated with companies, particularly their sustainability reports and descriptions, to predict corporate sustainability scores and classify companies accordingly. Unlike purely textual approaches, which treat each document and/or company as an independent entity, \name{} adopts a ``relational'' perspective. In fact, it models sustainability reports as nodes in a  graph, which we call a ``sustainability graph'', and the relationships between sustainability reports and/or the companies to which they refer as edges. These edges are constructed based on different similarity or membership criteria.

Let $R = \{r_1, r_2, \ldots, r_n\}$ be the set of sustainability reports into consideration, where each report $r_i \in R$ is associated with a specific company $\zeta_i$ and a given reference year. $r_i$ is also associated with a label $y_i$, which corresponds to the sustainability metric to be predicted. This metric can be the overall ESG score or one of its components (Environmental, Social or Governance). Thus, \name{} is not limited to a single predictive target; rather, it can be applied to various ESG assessment dimensions. $R$ is partitioned into a training set $R_{train}$ and a test set $R_{test}$ applying a 80/20 split.

\begin{figure*}[ht!]
    \centering
    \includegraphics[width=\textwidth]{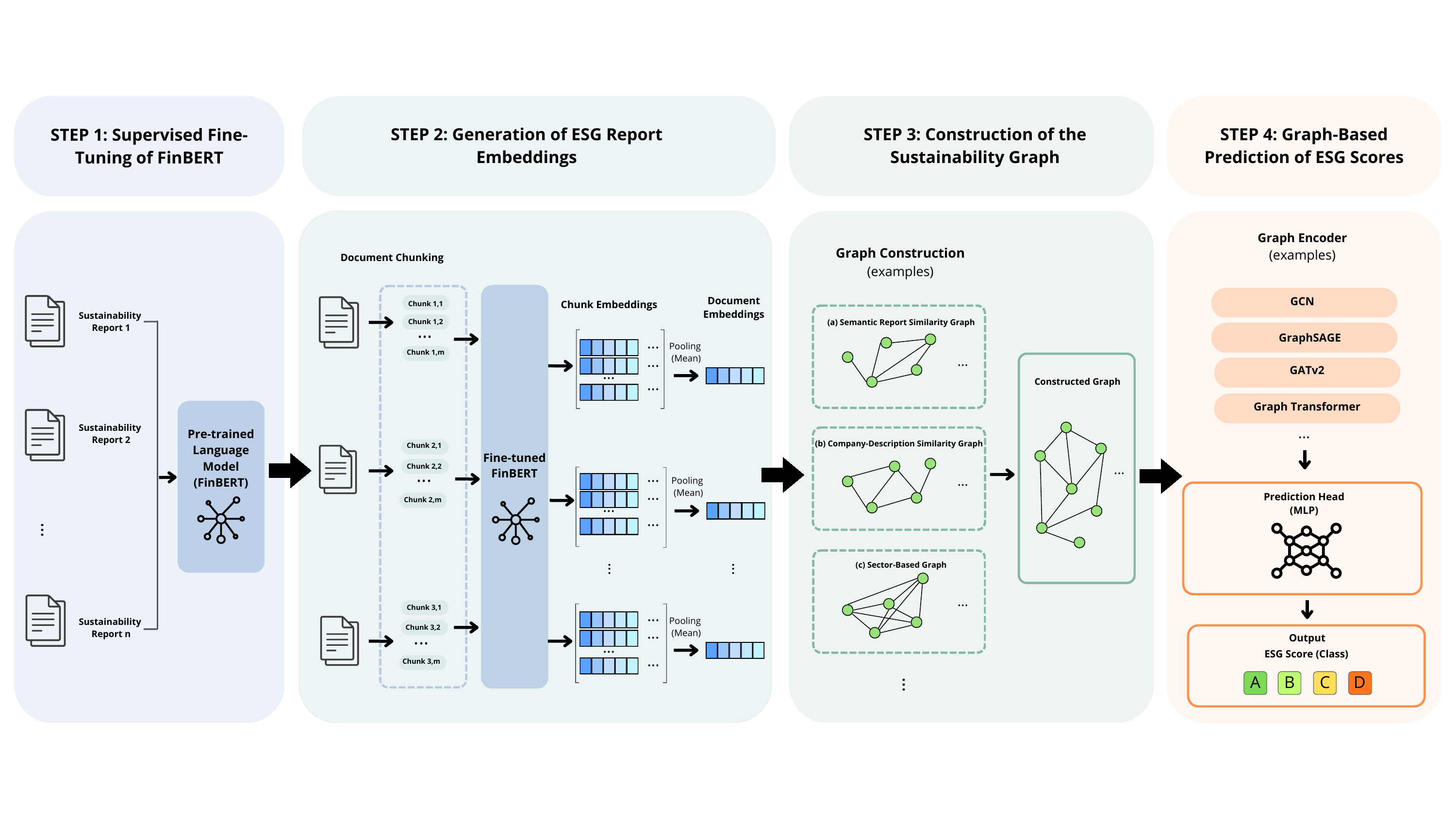}
    \caption{Overview of the \name{}'s workflow}
    \label{fig:framework}
\end{figure*}

Figure~\ref{fig:framework} illustrates the \name{}'s overall workflow, from processing sustainability reports to  constructing sustainability graphs and predicting the final ESG scores. As can be seen from this figure, \name{}'s  workflow consists of four steps.

In the first step, \name{} uses the sustainability reports to perform supervised fine-tuning of FinBERT, a Transformer model specialized for the financial domain \cite{Liu*21-7}. This step aims to adapt FinBERT to the corporate sustainability domain and to allow it to learn semantic representations more closely aligned with ESG metrics. Since sustainability reports are generally long and often exceed the maximum token limit supported by models, the text is segmented into chunks using dynamic or fixed segmentation strategies.

In the second step, \name{} uses the fine-tuned FinBERT to generate document-level embeddings. Specifically, \name{} divides each report into chunks and extracts a dense embedding for each chunk. Then, it averages the embeddings associated with the individual chunks to obtain a single embedding of the report.

In the third step, \name{} uses the obtained embeddings as initial features for the nodes of a sustainability graph $G = (V,E)$. Each node $v_i \in V$ represents the sustainability report of a company for a given year. The edges of $E$ represent relationships among the sustainability reports and/or the companies to which the sustainability reports refer. As previously pointed out, different kinds of relationship can be considered. Each of these kinds can give rise to a different sustainability graph.

In the fourth step, $G$ is used as input for a graph-based predictive model. This model combines the representation of each node with the representation of the nodes belonging to its neighborhood in the sustainability graph. Thus, the final prediction depends not only on the content of the report associated with a specific company, but also on the information provided by the neighboring reports and/or companies in the sustainability graphs. The latter are clearly different for the different kinds of company relationship taken into consideration. Therefore, this procedure ensures that ESG scores derive from both text information stored in the report and company relationships modeled in the sustainability graphs into consideration.

The \name{}'s key feature is its modularity. Each \name{}'s component (i.e., the model used to generate embeddings, the document segmentation strategy, the method for constructing sustainability graphs, the graph-based predictive model, and the target label) can be replaced or configured independently. 

In the following subsections, we illustrate in more detail the various steps of the \name{}'s workflow.

\subsection{Step 1: Supervised Fine-Tuning of FinBERT}
\label{sub:Step-1}

The first step of the \name{}'s workflow involves the supervised fine-tuning of FinBERT \cite{Liu*21-7} to adapt it to the domains of corporate sustainability and ESG assessment. Specifically, this fine-tuning aims to allow FinBERT to learn semantic representations of sustainability reports closely related to ESG classes. It also allows FinBERT to learn language representations already specialized in economic and financial domains. This improves the \name{}'s ability to interpret  terminology typically found in corporate reports and ESG documentation. \name{} performs the fine-tuning of FinBERT by providing it with the set $R_{train}$ of the training sustainability reports into consideration.

Due to their length, sustainability reports often exceed the 512-token maximum supported by the BERT architecture. To overcome this limitation, \name{} employs a text segmentation strategy that uses overlapping chunks. First, \name{} tokenizes the text of each report using the WordPiece tokenizer, which is compatible with FinBERT. At the end of this task, the resulting token sequence is divided into chunks with a maximum length of 510 tokens. The special tokens \texttt{[CLS]} and \texttt{[SEP]} are then added to these chunks, producing final chunks of 512 tokens. To maintain semantic continuity between consecutive chunks, \name{} uses a sliding window with a stride of 256 tokens, which introduces controlled overlap between adjacent chunks.

To address class imbalance, which is a significant issue in the ESG context, \name{} uses Weighted Focal Loss \cite{Lin*17-1} during the fine-tuning of FinBERT. Focal Loss reduces the contribution of easily classifiable samples, thereby focusing the learning process on more difficult samples. This improves the model's robustness with respect to minority classes. The Focal Loss used in \name{} is formally defined as follows:

\begin{equation}
\mathcal{L}_{FL} = - \alpha_t (1-p_t)^\gamma \log(p_t)
\end{equation}

Here, $p_t$ is the predicted probability of the correct class; $\alpha_t$ is the weight associated with the class $t$, which is inversely proportional to the class's relative frequency in the training set; $\gamma$ is the focusing parameter that controls the attention given to difficult samples.

\name{} trains FinBERT in supervised mode using the training set; its final configuration is determined by optimizing the F1-Score on the validation set. To prevent overfitting, \name{} employs an early stopping strategy based on the trend of the F1-Score.

The fine-tuned FinBERT is able to generate document embeddings semantically specialized for the ESG domain and is used by \name{} for constructing the embeddings of sustainability reports.

\subsection{Step 2: Generation of ESG report embeddings}
\label{sub:Step-2}

The second step of \name{}'s workflow involves generating dense vector representations of sustainability reports, which are then used as initial features for sustainability graph nodes. The goal of this step is to obtain document-level embeddings that can semantically represent the content of sustainability reports preserving information even when it is stored in particularly long documents.

As discussed in Section \ref{sub:Step-1}, sustainability reports often exceed the maximum token limit supported by Transformer architectures. For this reason, each report is divided into chunks compatible with the language model being used (e.g., 512-token chunks compatible with FinBERT). \name{} supports two segmentation strategies, i.e., dynamic and fixed. In the case of dynamic splitting, the number of chunks increases with the report length. Specifically, the report is divided into chunks, each of which has a length equal to the Transformer's capacity. This is achieved using a sliding window with controlled overlap between consecutive chunks. This strategy preserves the report's semantic structure and automatically adjusts segmentation according to the report's actual length. In the case of fixed splitting, each report is divided into a predefined number of chunks, regardless of its total length. We considered different values for the number of chunks to analyze the impact of text granularity on the quality of the generated embeddings.

Formally, given a report $r_i \in R$, the segmentation process produces a set of chunks:

\begin{equation}
C_i = \{c_{i_1}, c_{i_2}, \ldots, c_{i_m}\}
\end{equation}

\name{} supports various embedding models, albeit the default model is the fine-tuned FinBERT obtained as described in Section \ref{sub:Step-1}. Thus, \name{}'s behavior is independent of the linguistic model adopted for performing chunk embeddings. For each chunk $c_{i_j}$, \name{} obtains a dense embedding $\mathbf{z}_{i_j} \in \mathbb{R}^{d}$, where $d$ represents the dimensionality of the linguistic model's latent space. To obtain a single vector representation of the complete report, \name{} aggregates the embeddings of the individual chunks using a mean pooling operation. Formally, the final embedding associated with the report $r_i$ is obtained as follows:

\begin{equation}
\mathbf{z}_i =
\frac{1}{m}
\sum_{j=1}^{m}
\mathbf{z}_{i_j}
\end{equation}

This aggregation strategy generates stable, semantically meaningful document-level representations by combining the contributions of a report's chunks into a single, dense vector. These embeddings are then used as initial features for graph nodes when constructing the sustainability graph (see Section~\ref{sub:Step-3}).

\subsubsection{Possible embedding models to integrate in \name{}}
\label{subsub:Embedding-Models}

As previously pointed out, \name{} is agnostic with respect to the embedding model used internally. In our experiments, we analyzed two models for generating embeddings of sustainability reports. They are:

\begin{itemize}

\item {\em Fine-Tuned FinBERT} \cite{Liu*21-7}. This model is fine-tuned for the supervised ESG classification task as described in Section \ref{sub:Step-1}. It is the default model in \name{}. 

\item {\em Fine-Tuned Jina Embeddings (v2)} \cite{Gunter*23}. The Jina model family was designed to generate high-dimensional, general-purpose semantic embeddings. Unlike FinBERT, Jina was not specifically trained on the financial  domain. Analogously to FinBERT, we fine-tuned Jina for the supervised ESG classification task.

\end{itemize}

\subsection{Step 3: Construction of the graph}
\label{sub:Step-3}

The third step of the \name{}'s workflow creates a sustainability graph that models the dependencies among the considered sustainability reports (see Section \ref{sec:Proposed-Approach}). Given a set $R$ of sustainability reports and a kind of relationship $\varrho$ of interest, the sustainability graph $G_\varrho$ can be modeled as:

\begin{equation}
G_\varrho = (V_\varrho,E_\varrho)
\label{eq:G}
\end{equation}

 Each node $v_i \in V_\varrho$ corresponds to a sustainability report $r_i \in R$; therefore, the number of nodes in $G_\varrho$ coincides with the number of reports in $R$. Each node $v_i$ has associated a feature vector $\mathbf{z}_i \in \mathbb{R}^d$, which is the embedding of $r_i$ obtained in the previous step. 

Given a node $v_i \in V_\varrho$, we indicate by ${\cal N}(v_i)$ the set of neighbors of $v_i$, i.e., the set of nodes in $G_\varrho$ directly connected to $v_i$.

The most important part of this step is defining the kinds of relationship of interest. \name{} supports different kinds of relationship on the same set of sustainability reports. Each kind of relationship models a specific type of correlation between companies or reports. For example, one kind of relationship considers the semantic similarity of reports' textual content, another one the similarity of descriptive information associated with companies, and another one structural or categorical relationships between companies (e.g., the fact that they belong to the same industrial sector).

This formulation allows us to model both latent relationships derived from the linguistic content of reports and explicit relationships associated with organizational or structural characteristics of the corresponding companies. Generally, an edge between two nodes indicates that the corresponding nodes share a specific form of proximity, as defined by the kind of relationship which it refers to. Formally, each edge $e_{ik} \in E_\varrho$ connects two nodes $v_i$ and $v_k$ and can be associated with a weight $w_{ik} \in \mathbb{R}$, representing the strength of the relationship between the reports represented by $v_i$ and $v_k$ or the companies to which they refer. This weight can then be used in the prediction of ESG scores.

The \name{}'s capability of simultaneously handling different kinds of relationship enables a systematic analysis of how different kinds of relationship affect the quality of learned representations and the final performance of the ESG prediction task.

\subsubsection{Graph Construction Strategies}
\label{subsub:Graph-Construction-Strategies}

As we have seen above, \name{} is agnostic with respect to the kind of relationship associated with the sustainability graph $G_\varrho$ (see Equation \ref{eq:G}). In our experiments, we considered latent correlations derived from the reports' content and explicit correlations associated with the companies' descriptive or structural characteristics. In the following we present three kinds of relationship. The first is an example of latent correlation; the second and the third are examples of explicit correlations.

\paragraph{Semantic Report Similarity Graph}
\ \\
The first kind of relationship used for constructing the sustainability graph is the semantic similarity of the text in sustainability reports. In this case, the edges of $G\varrho$ and the corresponding weights are directly derived within the embedding space at the document level. Let $\mathbf{z}_i$ and $\mathbf{z}_k$ be the embeddings associated with the reports $r_i$ and $r_k$, respectively. The similarity between $r_i$ and $r_k$ is obtained by calculating the cosine similarity between $\mathbf{z}_i$ and $\mathbf{z}_k$:

\begin{equation}
{\tt sim}(r_i,r_k) =
\frac{
\mathbf{z}_i \cdot \mathbf{z}_k
}{
\|\mathbf{z}_i\| \|\mathbf{z}_k\|
}
\end{equation}

The existence of an edge between the nodes $v_i$ and $v_k$ in $G_\varrho$, corresponding to the reports $r_i$ and $r_k$, depends on whether the value of ${\tt sim}(r_i,r_k)$ is greater than or equal to a similarity threshold $\tau_{cos}$:

\begin{equation}
e_{ik} \in E_\varrho
\iff
{\tt sim}(r_i,r_k) \geq \tau_{cos}
\end{equation}

This allows $G_\varrho$ to link reports with semantically related content and enables graph-based prediction models to consider the existence of similarities between linguistic patterns, ESG terminology, or narrative structures. The resulting cosine similarity is used as the weight $w_{ik}$ associated with $e_{ik}$.

\paragraph{Company Description Similarity Graph}
\ \\
The second kind of relationship used for constructing the edges of $G_\varrho$ is the similarity of the descriptive information about companies associated with reports. In this case, the edges of $G_\varrho$ model descriptive and organizational-semantic similarities. The nodes of $G_\varrho$ are associated with sustainability reports, while edges are constructed using information extracted automatically from company descriptions via Natural Language Processing (NLP) techniques. Specifically, the SpaCy library is used to perform a text extraction pipeline consisting of two complementary modules devoted to performing morphosyntactic analysis and Named Entity Recognition (NER), respectively.

Morphosyntactic analysis identifies semantically relevant nouns and proper nouns within company descriptions. The extracted terms are then lemmatized and filtered to reduce linguistic variability and standardize semantically equivalent words. NER  identifies named entities belonging to specific categories of information, including: {\em (i)} organizations and companies (\texttt{ORG}), {\em (ii)} geopolitical entities (\texttt{GPE}), {\em (iii)} locations (\texttt{LOC}), {\em (iv)} people (\texttt{PERSON}), {\em (v)} products, brands, and technologies (\texttt{PRODUCT}). Combining these two techniques overcomes the limitations of each one. In fact, morphosyntactic analysis captures informative keywords but can break down complex expressions, while NER preserves complete semantic entities but tends to ignore relevant common terms.

At the end of the process, the extracted keywords and entities related to the company $\zeta_i$ associated with a report $r_i$ are merged into one set:

\begin{equation}
S_i = \{s_{i_1}, s_{i_2}, \ldots, s_{i_m}\}
\end{equation}

$S_i$ represents a description of $\zeta_i$. The similarity between the sets $S_i$ and $S_k$, relative to the companies $\zeta_i$ and $\zeta_k$ associated with the reports $r_i$ and $r_k$, is calculated using the Jaccard coefficient between $S_i$ and $S_k$:

\begin{equation}
J(S_i,S_k) =
\frac{
|S_i \cap S_k|
}{
|S_i \cup S_k|
}
\end{equation}

The presence of an edge $e_{ik}$ between the two nodes $v_i$ and $v_k$ depends on whether $J(S_i,S_k)$ is greater than or equal to a threshold $\tau_J$:

\begin{equation}
e_{ik} \in E_\varrho
\iff
J(S_i,S_k) \geq \tau_J
\end{equation}

Also in this case, the similarity value is used as the weight $w_{ik}$ of the edge $e_{ik}$. This configuration allows for the creation of a sustainability graph based on descriptive similarities among companies, regardless of the direct similarity of their sustainability reports.

\paragraph{Sector-Based Graph}
\ \\
The third kind of relationship used for constructing the edges of $G_\varrho$ is based on the industry classification of the companies to which the reports refer. Formally, let $r_i$ and $r_k$ be two reports and let $\zeta_i$ and $\zeta_k$ be the companies to which they refer. Let $v_i$ and $v_k$ be the nodes in $G_\varrho$ associated with $r_i$ and $r_k$, respectively. Let ${\tt sector}(\zeta_i)$ be the function that returns the industrial sector associated with $\zeta_i$. The presence of an edge between $v_i$ and $v_k$ depends on whether the industrial sectors of $\zeta_i$ and $\zeta_k$ are the same:

\begin{equation}
e_{ik} \in E_\varrho
\iff
{\tt sector}(\zeta_i)
=
{\tt sector}(\zeta_k)
\end{equation}

In this case, the weight of all the edges of $G_\varrho$ thus obtained is set to 1. This procedure generates a sustainability graph consisting of multiple connected components, each associated with a specific industrial sector. The underlying assumption is that companies within the same industrial sector have similar operational practices, regulatory constraints, ESG strategies, and industry terminology.\\

This variety of relationship types that can be considered for constructing sustainability graphs allows us to systematically evaluate how different forms of correlation between reports and/or companies contribute to ESG predictions. Since the nodes of the sustainability graphs are the same across the different kinds of relationship, observed differences in performance are primarily due to effects of the modeled kinds of relationship.

\begin{figure}[ht!]
    \centering
    % Prima immagine
    \begin{subfigure}[t]{0.49\textwidth}
        \centering
        \includegraphics[width=\textwidth]{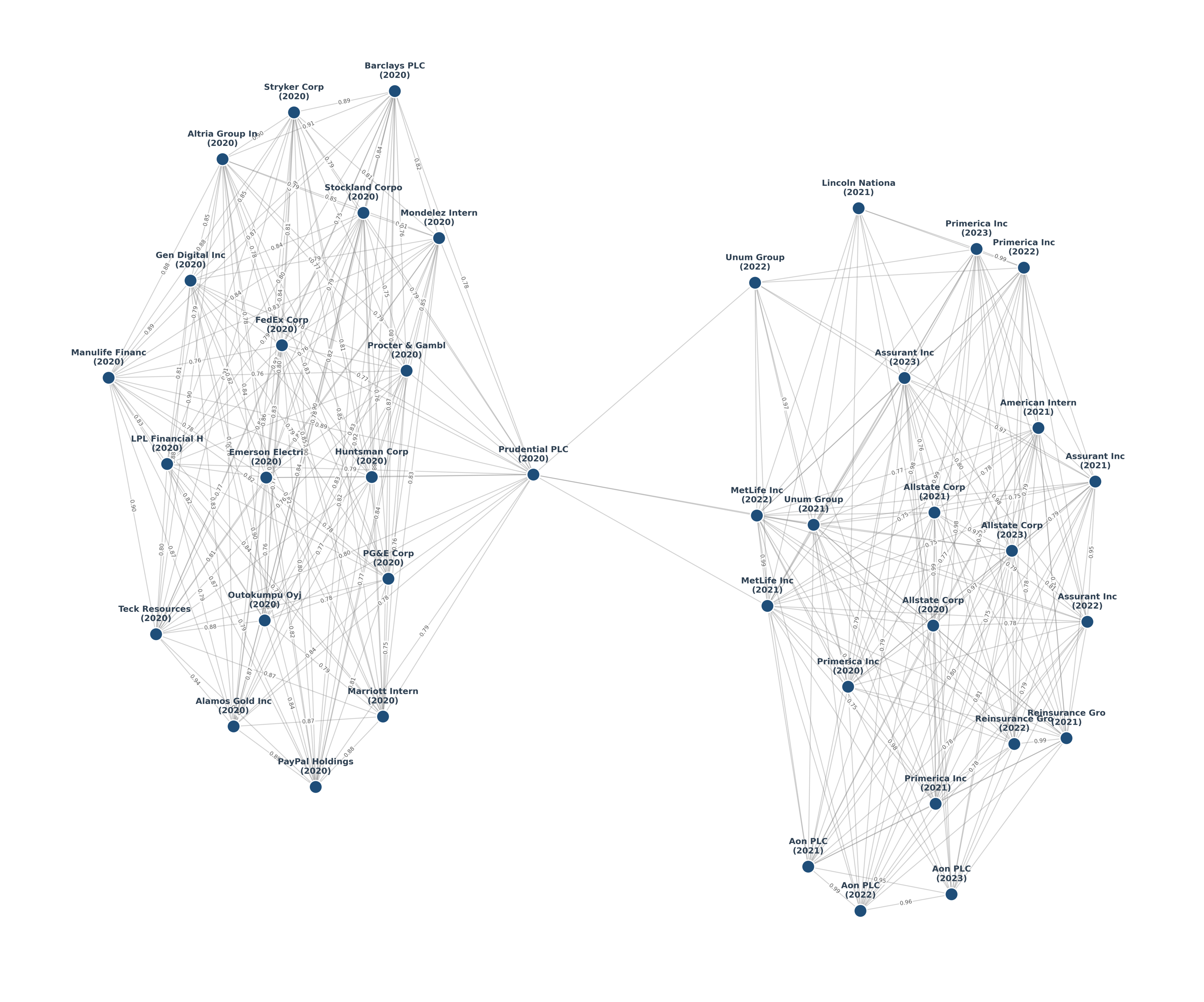}
        \caption{Semantic Report Similarity Graph}
    \end{subfigure}
    \hfill
    % Seconda immagine
    \begin{subfigure}[t]{0.49\textwidth}
        \centering
        \includegraphics[width=\textwidth]{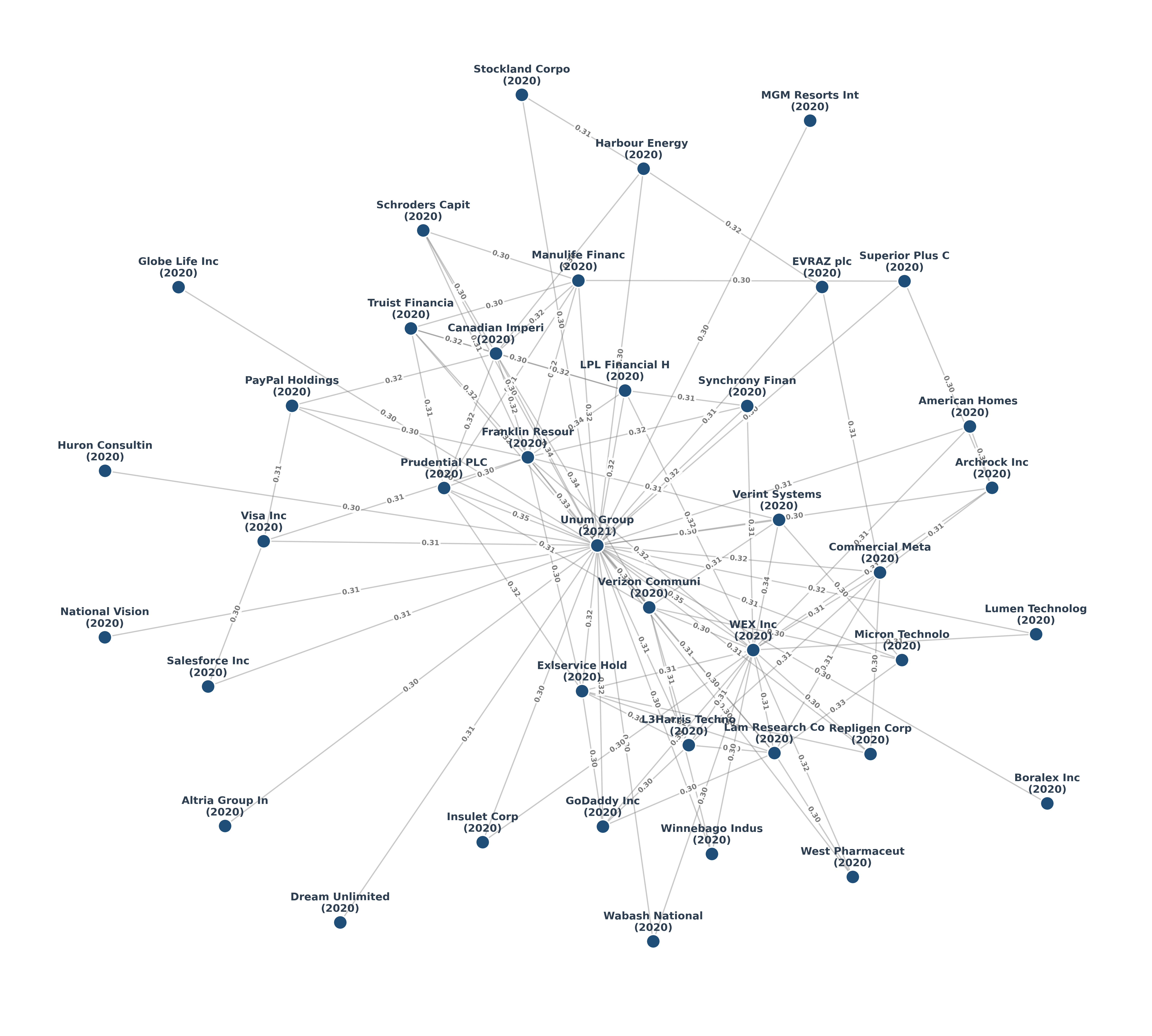}
        \caption{Company Description Similarity Graph}
    \end{subfigure}
    \hfill
    % Terza immagine
    \begin{subfigure}[t]{0.49\textwidth}
        \centering
        \includegraphics[width=\textwidth]{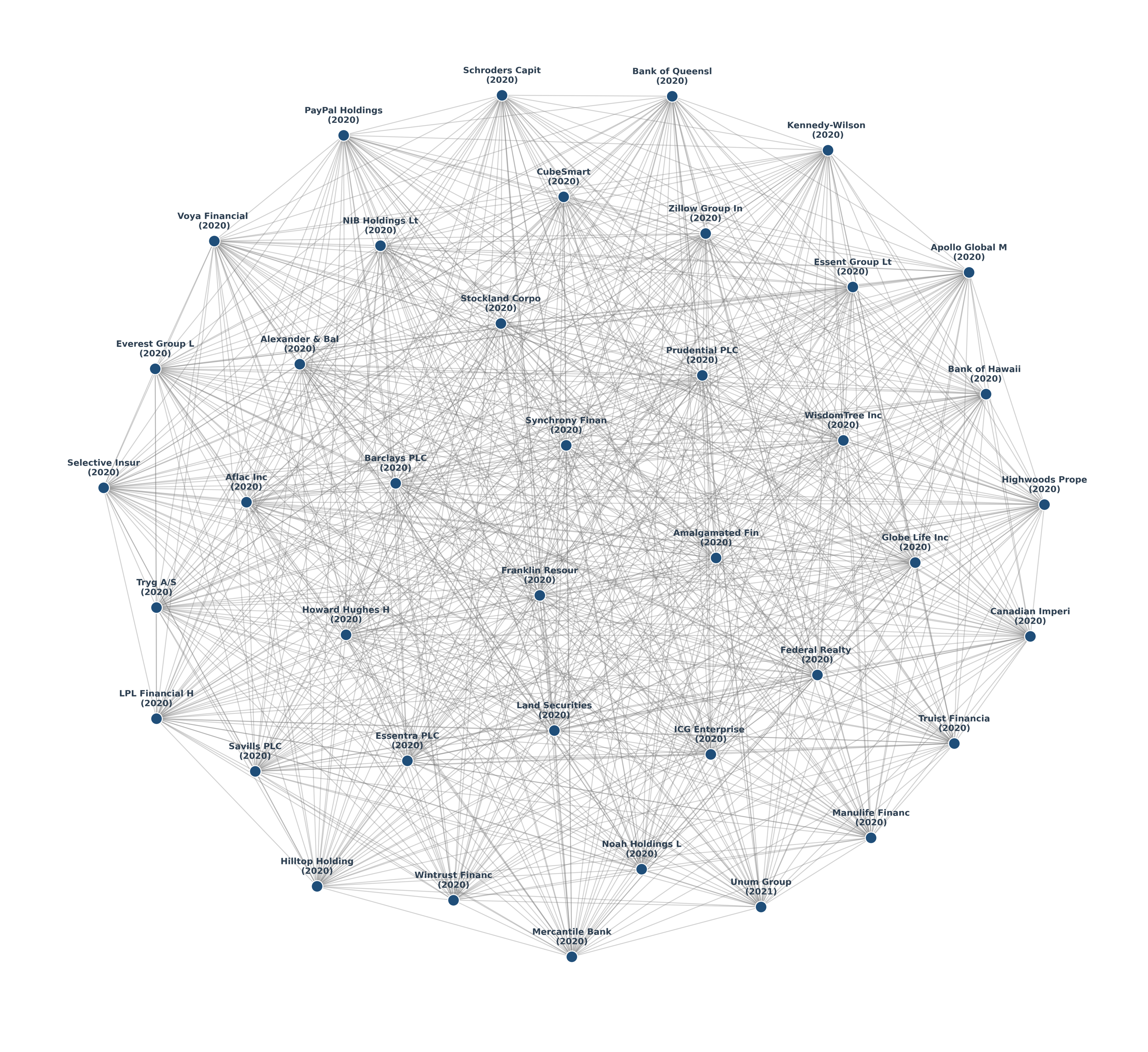}
        \caption{Sector-Based Graph}
    \end{subfigure}
    
    \caption{Sustainability graphs obtained by applying Semantic Report Similarity Graph, Company Description Similarity Graph and Sector-Based Graph on the same nodes}
    \label{fig:confronto_grafi}
\end{figure}

Figure~\ref{fig:confronto_grafi} illustrates an example of the three graph construction strategies described above. In particular, the three subfigures show three sustainability graphs, each consisting of 40 nodes representing 40 sustainability reports relative to companies belonging to the same industrial sector. The first (resp., second, third) subfigure shows the Semantic Report Similarity Graph (resp., Company Description Similarity Graph, Sector-Based Graph) obtained from the same reports. The figure highlights the markedly different edge distributions produced by the three graph construction strategies. The Semantic Report Similarity Graph forms relatively compact clusters of companies with similar sustainability reports. Instead, the Company Description Similarity Graph exhibits a more heterogeneous connectivity pattern, reflecting similarities in the organizational characteristics of companies. Finally, the Sector-Based Graph is considerably denser because it connects companies belonging to the same industrial sector.

%%%%%%%%%
%%%%%%%%%

% Figura dei grafi!

%%%%%%%%%
%%%%%%%%%

\subsection{Step 4: Graph-Based Prediction of ESG Scores}
\label{sub:Step-4}

The fourth and final step of the \name{}'s workflow employs graph-based predictive models to predict the ESG scores associated with sustainability reports and to classify them accordingly. Starting from the graph $G_\varrho=(V_\varrho,E_\varrho)$ constructed in the previous step, \name{} learns contextualized representations of nodes by considering the relationships linking the nodes of $G_\varrho$ and the corresponding strengths expressed by edge weights. Therefore, the predicted class of a report $r_i$ corresponding to a node $v_i$ in $G_\varrho$ depends not only on the textual content of $r_i$, but also on the information derived from the nodes belonging to the neighborhood ${\cal N}(v_i)$ of $v_i$ in $G_\varrho$. Clearly, ${\cal N}(v_i)$ depends on the edges of $G_\varrho$ (see Section \ref{sec:Proposed-Approach}), and therefore on the relationship type $\varrho$ associated with $G_\varrho$.

Formally, let $\mathbf{Z} \in \mathbb{R}^{n \times d}$ be the matrix of the embeddings of the reports associated with the nodes of $G_\varrho$, where the row $\mathbf{z}_i$ corresponds to the embedding of the report $r_i$. A graph-based predictive model adopted by \name{} learns a parametric function:

\begin{equation}
f_\theta : (G_\varrho,\mathbf{Z}) \rightarrow \hat{\mathbf{Y}}
\end{equation}

Here, $\theta$ is the set of learnable parameters and $\hat{\mathbf{Y}}$ is the matrix of predictions generated for the nodes of $G_\varrho$, and therefore for the reports of $R$. Here, we assume that the graph-based predictive model consists of $L$ layers. In general, each layer updates the representation of a node $v_i$ by aggregating information from its neighborhood ${\cal N}(v_i)$. Let $\mathbf{h}_i^{0} = \mathbf{z}_i$ denote the initial representation of $v_i$. The representation of $v_i$ updated at the layer $\ell+1$ can be expressed as follows:

\begin{equation}
\mathbf{h}_i^{\ell+1} =
\phi^{\ell}
\left(
\mathbf{h}_i^{\ell},
{\tt AGG}
\left(
\left\{
\mathbf{h}_k^{\ell}, e_{ik}, w_{ik}
\,:\,
v_k \in \mathcal{N}(v_i)
\right\}
\right)
\right)
\end{equation}

Here: {\em (i)} ${\tt AGG}$ is an aggregation function depending on the adopted prediction model; {\em (ii)} $\phi^{\ell}$ is a learnable transformation that may be combined with nonlinear activation functions, normalization, and regularization.

\name{} can support various graph-based predictive models, including Graph Convolutional Networks, GraphSAGE, Graph Attention Networks, and Graph Transformers. These models differ primarily in how they aggregate and weight information from adjacent nodes. Some approaches use normalized aggregations based on the graph structure. Others learn neighborhood sampling and aggregation functions. Still others use attention mechanisms to assign varying importance to the nodes of the neighborhood or the edges connecting them.

Let $\mathbf{h}_i^{L}$ be the representation of the node $v_i$ in the final layer of the graph-based predictive model. \name{} inputs $\mathbf{h}_i^{L}$ into a final classifier $\Gamma$, which returns the probability distribution associated with all the possible ESG classes. The class with the highest probability is returned as the class of $r_i$:

\begin{equation}
\hat{y}_i =
{\tt argmax}(\Gamma(\mathbf{h}_i^{L}))
\end{equation}

The graph-based predictive model is trained in a supervised manner using the node labels available in the training set. The loss is computed by comparing the predicted class $\hat{y}^i$ with the corresponding ground-truth label $y_i$. To mitigate the effects of class imbalance, ESGPred employs weighted loss functions, such as Weighted Cross-Entropy or Focal Loss, thereby reducing the bias toward majority classes and improving predictive performance on underrepresented classes.

Due to its modular design, ESGPred can be adapted to predict different types of sustainability-related information while preserving its overall architecture and workflow. Specifically, the predictive graph-based model underlying \name{} can be trained to predict various sustainability metrics. The first metric is the overall ESG score, which is an aggregate measure of corporate performance across the Environmental, Social, and Governance dimensions. In this case, the label $y_i^{ESG}$ associated with the report $r_i$ indicates the company's overall ESG class. In addition to the overall ESG score, \name{} can predict the individual ESG components. For the Environmental component, the label $y_i^{E}$ represents the company's assessment with respect to issues such as environmental impact, emissions, resource use, energy management, and climate policies. For the Social component, the label $y_i^{S}$ reflects aspects related to human resources, working conditions, community relationships, human rights, safety, and social responsibility. Finally, for the Governance component, the label $y_i^{G}$ refers to elements such as board structure, oversight practices, transparency, corporate ethics, compensation, and risk management.

Formally, \name{} allows a set of possible prediction targets:

\begin{equation}
\mathcal{T} = \{ESG, E, S, G\}
\end{equation}

For each target $t \in \mathcal{T}$, \name{} can associate a set of labels $\mathbf{Y}^{t} = \{y_1^{t}, y_2^{t}, \ldots, y_n^{t}\}$ such that $y_i^t$, $1 \leq t \leq n$, is the score of the report $r_i \in R$. The graph-based predictive model underlying \name{} is trained to learn a function specific to each target:

\begin{equation}
f_\theta^{t} : (G,\mathbf{Z}) \rightarrow \hat{\mathbf{Y}}^{t}
\end{equation}

\subsubsection{Graph-Based Predictive Models}
\label{subsub:Graph-Based-Predictive-Models}

\name{} is agnostic with respect to the graph-based predictive model employed. In our experiments, we considered various Graph Neural Networks (GNNs) as predictive models to analyze the impact of different aggregation, propagation, and attention mechanisms on the quality of the provided predictions. In all cases, each node in the sustainability graph is associated with an initial feature vector:

\begin{equation}
\mathbf{h}_i^{0} = \mathbf{z}_i,
\end{equation}

Here, $\mathbf{z}_i \in \mathbb{R}^{d}$ is the embedding of the report $r_i$. The prediction activity consists of iteratively updating these representations through message-passing operations between connected nodes of the sustainability graph.

The predictive models under consideration differ primarily in the following aspects: \emph{(i)} the strategy for aggregating information from the neighborhood; \emph{(ii)} the weights of the contributions from neighborhing nodes of the sustainability graph; \emph{(iii)} the explicit use of edge weights; \emph{(iv)} the mechanism for propagating information throughout the sustainability graph. They are:

\begin{itemize}

\item {\em Graph Convolutional Network} (GCN) \cite{KiWe16-1}. It uses approximate spectral convolution to propagate information between adjacent nodes. Instead of averaging the values of neighboring nodes, GCN applies a symmetric normalization based on node degrees, allowing the contribution of connected nodes to be adjusted according to the local graph structure. In \name{}, the GCN model explicitly uses the weights associated with the sustainability graph's edges. 

\item {\em GraphSAGE} \cite{HaYiLe17}. Unlike traditional GCNs, GraphSAGE employs parameterized neighborhood aggregation schemes, enabling a more flexible and expressive integration of local neighborhood information.

\item {\em Graph Attention Network} (GAT) \cite{Velickovic*18}. It is based on attention mechanisms applied to node neighborhoods. In this model, the contribution of adjacent nodes is not solely determined by the sustainability graph structure; rather, it is learned dynamically through attention coefficients. 

\item {\em Graph Transformer} \cite{DiBr20}. It combines Transformer principles with the message passing characteristic of Graph Neural Networks. Unlike previous architectures, the Graph Transformer uses multi-head attention mechanisms to simultaneously model different kinds of relationship between the sustainability graph nodes. Specifically, each attention head learns a different neighborhood weighting function, allowing the predictive model to capture multiple, potentially complementary relationships. 

\end{itemize}

\section{Experiments}
\label{sec:Experiments}

In this section, we describe our experiments conducted to evaluate \name{}. In particular, Subsection \ref{sub:Experimental-Setup} presents the experimental setup. Subsection \ref{sub:results} reports our experimental results and compares \name{} with several related approaches. Finally, Subsection \ref{sub:ablation} presents a study of \name{}'s hyperparameters.

\subsection{Experimental Setup}
\label{sub:Experimental-Setup}

In this section, we illustrate the datasets, baselines and metrics adopted in our experiments.

\subsubsection{Dataset}
\label{subsub:dataset}

For our experiments, we downloaded sustainability reports in PDF format from   \emph{ResponsibilityReports}\footnote{\url{https://www.responsibilityreports.com/}}, a platform that aggregates official company documentation. In addition to the reports, the platform provides complementary metadata for each company, including sector classification, country of origin, and a descriptive profile. These descriptions range from brief summaries spanning a few lines to more detailed accounts spanning several paragraphs. They typically address the company's industry affiliation, founding history, geographical roots, core business activities, and principal products or services it provides. They may also convey the company's mission or values. Our collection covers the period from 2020 to 2023 and is restricted to public companies. This selection is based on two practical considerations: {\em (i)} documents from public companies adhere to more standardized disclosure norms and are consistently accessible, and {\em (ii)} ground-truth ESG scores from our reference provider are available for this category of companies. While the raw collection contains 10,014 reports, our final experimental dataset focuses on a subset of them. The sectors most represented in the final sample are: finance (519 reports), technology (486), industrial goods (461), basic materials (428), and consumer goods (419). Conversely, the countries with the most reports are the United States (1,201), England (234), and Australia (212). The overall geographic distribution of companies to which sustainability reports refer is shown in Table \ref{tab:Full-Geographic-Distribution}.

\begin{table}[ht!]
\centering
{\scriptsize
\begin{tabular}{ll|ll}
\toprule
\textbf{Country} & \textbf{Count} & \textbf{Country} & \textbf{Count} \\
\midrule
USA & 1,201 & Luxembourg & 3 \\
England & 234 & Spain & 2 \\
Australia & 212 & Belgium & 2 \\
Canada & 205 & Chile & 2 \\
Sweden & 52 & Hong Kong & 2 \\
Finland & 23 & Cayman Islands & 2 \\
Denmark & 20 & Italy & 2 \\
Ireland & 18 & Wales & 2 \\
China & 15 & Argentina & 2\\
Bermuda & 15 & Cyprus & 1 \\
Greece & 14 & Austria & 1 \\
Switzerland & 12 & Isle of Man & 1 \\
Singapore & 10 & Peru & 1 \\
Scotland & 9 & Panama & 1 \\
Israel & 9 & Kazakhstan & 1 \\
Germany & 8 & Mauritius & 1 \\
Guernsey & 7 & Ghana & 1  \\
Jersey & 6 & Mexico & 1 \\
France & 4 & Lithuania & 1 \\
Brazil & 4 & Papua New Guinea & 1 \\
Netherlands & 4 & Slovenia & 1 \\
New Zealand & 4 & Taiwan & 1 \\
India & 3 & Turkey & 1 \\
South Africa & 3 & Thailand & 1 \\
Monaco & 3 & Ukraine & 1 \\
Nigeria & 3 & Uruguay & 1 \\
\midrule
{\bf Total} & & 2,134 \\
\bottomrule
\end{tabular}
}
\caption{Overall geographical distribution of the companies to which sustainability reports refer}
\label{tab:Full-Geographic-Distribution}
\end{table}

All reports underwent a structured preprocessing pipeline before use. This pipeline was designed to convert PDF content into plain text. Non-textual elements, including figures, charts, and graphical layouts, were systematically discarded. The pipeline proceeded through three stages: {\em (i)} raw text extraction from the source PDF files, {\em (ii)} normalization of character encoding and removal of formatting residues, and {\em (iii)} filtering of non-informative recurring elements, such as page numbers and repeated section headers.

The preprocessed corpus, organized as one text document per company per fiscal year, served two purposes in \name{}. First, it provided the training set for fine-tuning FinBERT. Second, it supplied the embeddings used to construct the sustainability graph, which is the input for graph-based predictive models.

To validate \name{}, we gathered ESG scores for public companies. Specifically, we used the scores provided by ``LSEG Data \& Analytics'' (formerly Refinitiv) via the ``LSEG Workspace'' platform\footnote{\url{https://www.lseg.com/en/data-analytics/products/workspace}}. These scores summarize a company's overall ESG standing by combining different indicators through a proprietary aggregation methodology.

We compared the ESG scores generated by \name{} with the corresponding scores provided by LSEG. To ensure a valid comparison, we restricted the dataset to companies for which both the sustainability report and the LSEG ESG score were available for the same reference year.

As LSEG ESG data is subject to a commercial license that prohibits public redistribution, we cannot disclose individual scores in this paper. Instead, we report experimental results in terms of aggregate statistics and score quartiles encoded as letter grades ranging from $A$ to $D$, where $A$ indicates the highest-performing quartile and $D$ denotes the lowest one.

Table \ref{tab:Benchmark-Statistics} reports some statistics related to our dataset. This dataset corresponds to the set $R$ of sustainability reports mentioned in Section \ref{sec:Proposed-Approach}. As pointed out in that section, we divided $R$ into a training set $R_{train}$ and a test set $R_{test}$, applying a 80/20 split. We assigned all reports associated with the same company exclusively to either the training set or the test set to prevent data leakage.

\begin{table}[h!]
\centering
{\scriptsize
\begin{tabular}{ll}
\toprule
{\em Name} & {\em Value}  \\
\midrule
Total number of sustainability reports & 5,266 \\
Total number of reports relating to 2020 & 1,133 \\
Total number of reports relating to 2021 & 1,459 \\
Total number of reports relating to 2022 & 1,665 \\
Total number of reports relating to 2023 & 1,009 \\
Total number of reports with score $A$ & 785 \\
Total number of reports with score $B$ & 2,652 \\
Total number of reports with score $C$ & 1,611 \\
Total number of reports with score $D$ & 218 \\
Total number of companies & 2,134 \\
Total number of countries of origin of companies & 52 \\
\bottomrule
\end{tabular}
}
\caption{Some statistics of the dataset $R$ of reports used in our experiments}
\label{tab:Benchmark-Statistics}
\end{table}

\subsubsection{Baselines}
\label{subsub:baselines}

To evaluate the effectiveness of \name{}, we compared the results returned by it to three baselines representing the current state-of-the-art in text-based ESG classification. These baselines allow us to analyze the contributions of the various \name{}'s components and distinguish the effects of the language model's domain-specific specialization from the effects of the sustainability graph and graph-based predictive models. They are:

\begin{itemize}

\item {\em Vanilla FinBERT} \cite{Liu*21-7} in its original version. It was pre-trained on financial corpora, but not adapted for the ESG domain. In this configuration, the model is used directly as a supervised classifier for sustainability reports. This baseline evaluates the predictive power of generic financial language representations without adaptation to a specific task.

\item {\em Fine-Tuned FinBERT}. It was fine-tuned using the ESG dataset employed in this paper. In this case, the model was trained following the procedure described in Section \ref{sub:Step-1} with ESG scores as the supervised targets. This baseline is particularly relevant because it uses the same textual information and language adaptation tasks as \name{}, but does not introduce any sustainability graph or graph-based predictive model. Therefore, comparing \name{} to this baseline allows us to directly quantify the contributions of our graph modeling present in \name{} compared to an approach based solely on the  textual content of sustainability reports.

\item {\em Fine-Tuned Transformer} \cite{PaEh22}. It was originally employed as a part of an ESG classification system that used fine-tuned Transformer models, SEC Form 10-K regulatory documents, and ESG ratings provided by S\&P Global.  In the original paper, the problem is formulated as a binary classification task, which is achieved through a median split of ESG ratings. To ensure a fair comparison with \name{}, we reimplemented the approach described in \cite{PaEh22} to make it applicable to the dataset used in this paper. Specifically, we adapted it to a four-class classification problem (with the classes being $A$, $B$, $C$, and $D$), which is consistent with the categorization of the ESG scores provided by LSEG. Thus, the results obtained from this baseline allow us to compare \name{} with a Transformer-based approach specifically designed for ESG assessment and widely recognized in recent literature.

\end{itemize}

\subsubsection{Metrics}
\label{subsub:Metrics}

To evaluate and compare \name{} with the selected baselines, we used four classification metrics widely adopted in the literature: Accuracy, Precision, Recall, and F1-Score.

Let $Cl= \{A,B,C,D \}$ denote the set of ESG classes and let $y_i \in Cl$ denote the actual class associated with the report $r_i \in R$. Let $\hat{y}_i \in Cl$ denote the class predicted by the model. 

Accuracy is defined as the proportion of sustainability reports for which the predicted and actual ESG classes coincide. It is defined as follows:

\begin{equation}
Accuracy = \frac{1}{N}
\sum_{i=1}^{N}
\mathbb{I}(y_i=\hat{y}_i),
\end{equation}

where $N$ is the total number of reports in the test set and $\mathbb{I}(\cdot)$ is the indicator function, which returns 1 when the actual and predicted classes coincide and 0 otherwise.

For each ESG class $cl \in Cl$, Precision is defined as the proportion of reports predicted to belong to $cl$ that actually belong to $cl$. It is defined as follows:

\begin{equation}
P_{cl}=
\frac{TP_{cl}}{TP_{cl}+FP_{cl}},
\end{equation}

where $TP_{cl}$ is the number of reports whose actual and predicted class is $cl$  and $FP_{cl}$ is the number of reports predicted as class $cl$ despite belonging to a different actual class.

For each ESG class $cl \in Cl$, Recall measures the proportion of reports that actually belong to $cl$ and have been correctly assigned to $cl$ by the model. It is defined as follows:

\begin{equation}
R_{cl}=
\frac{TP_{cl}}{TP_{cl}+FN_{cl}},
\end{equation}

where $FN_{cl}$ is the number of reports whose actual class is $cl$ but whose predicted class is different.

For each ESG class $cl \in Cl$, F1-Score is defined as the harmonic mean of Precision and Recall. It is defined as follows:

\begin{equation}
F_{cl}=
2 \cdot \frac{P_{cl} \cdot R_{cl}}{P_{cl}+R_{cl}}.
\end{equation}

Since there are four classes in ESG, the final Precision, Recall and F1-Score values considered in our experiments were obtained by averaging the corresponding parameter values for each class: 

\begin{equation}
Precision = \frac{1}{4} \sum_{cl \in Cl} P_{cl}, \hspace{0.5cm} Recall = \frac{1}{4} \sum_{cl \in Cl} R_{cl}, \hspace{0.5cm} F1\text{-}Score = \frac{1}{4} \sum_{cl \in Cl} F_{cl}
\end{equation}

\subsection{Results}
\label{sub:results}

We conducted all experiments on a server equipped with two NVIDIA Tesla T4 GPUs, each with 16GB of GPU memory, and an Intel Xeon CPU (Skylake-SP architecture) with 2 cores and 4 hardware threads.

Table \ref{tab:results-baseline} shows the results of comparing \name{} with the baselines described in Section \ref{subsub:baselines}. In this table and in the next ones, the highest value of the performance metric associated with each column is shown in bold.

\begin{table}[ht!]
\centering
\scriptsize
\begin{tabular}{@{}llcccc@{}}
\toprule
\textbf{Category} & \textbf{Model / Architecture} & \textbf{Accuracy} & \textbf{Precision} & \textbf{Recall} & \textbf{F1-Score} \\ \midrule

\textbf{Baselines}         & & & & \\ 
                           & Vanilla FinBERT                         & 0.72 & 0.69 & 0.60 & 0.63 \\
                           & Fine-Tuned FinBERT                      & 0.77 & 0.72 & 0.74 & 0.73 \\
                           & Fine-Tuned Transformer     & 0.76 & 0.74 & 0.65 & 0.68 \\ \midrule

% Gruppo Proposed 1
\textbf{\name{}}          & & & & & \\
\textit{Description-Based Graph} & Graph Transformer                 & \textbf{0.85} & \textbf{0.83} & \textbf{0.85} & \textbf{0.84} \\
                           & GraphSAGE                        & 0.84 & 0.80 & 0.84 & \textbf{0.84} \\
                           & GAT                              & 0.81 & 0.76 & 0.82 & 0.78 \\
                           & GCN                              & 0.77 & 0.71 & 0.74 & 0.73 \\ \addlinespace

% Gruppo Proposed 2
\textit{Semantic-Based Graph}     & Graph Transformer                 & 0.84 & 0.80 & \textbf{0.85} & 0.82 \\
                           & GraphSAGE                        & 0.82 & 0.80 & 0.82 & 0.81 \\
                           & GAT                              & 0.82 & 0.78 & \textbf{0.85} & 0.81 \\
                           & GCN                              & 0.81 & 0.78 & \textbf{0.85} & 0.81 \\ \addlinespace

% Gruppo Proposed 3
\textit{Sector-Based Graph}      & Graph Transformer                 & 0.81 & 0.76 &         0.79 & 0.77 \\
                           & GraphSAGE                        & 0.78 & 0.73 & 0.75 & 0.74 \\
                           & GAT                              & 0.79 & 0.73 & 0.77 & 0.75 \\
                           & GCN                              & 0.76 & 0.70 & 0.74 & 0.72 \\ \bottomrule
\end{tabular}
\caption{Comparison between baselines and \name{} with different graph-based configurations}
\label{tab:results-baseline}
\end{table}

As expected, the Fine-Tuned FinBERT significantly outperforms the Vanilla FinBERT. In fact, Accuracy increases by 6.94\%, while F1-Score rises by 15.87\%. These results confirm the importance of adapting the language model used to obtain report embeddings to the ESG domain and demonstrate that representations learned during fine-tuning capture information relevant to classifying sustainability reports.

The Fine-Tuned Transformer yields results similar to those of the Fine-Tuned FinBERT, achieving an Accuracy of 0.76. Although this method was designed for binary classification, its adaptation to a multi-class context returns satisfying results. This confirms the effectiveness of Transformer-based strategies for analyzing ESG documents. Nevertheless, its performance remains inferior to the best configurations of \name{}.

Analyzing the various sustainability graph construction strategies in \name{} clearly shows that the quality of the sustainability graph significantly affects the performance of \name{}. The configuration that yields the best results is the Description-Based Graph combined with the Graph Transformer, which achieves an Accuracy of 0.85 and an F1-Score of 0.84. This suggests that data extracted from company descriptions is an effective source of useful information for ESG prediction. In fact, descriptions tend to capture the structural characteristics of a company, such as its industrial sector, products, technologies, target markets, and business type. This generates connections between companies that share similar economic and organizational features.

The Semantic-Based Graph produces slightly inferior but competitive results. The optimal configuration achieves an Accuracy of 0.84 and an F1-Score of 0.82. These results highlight the importance of semantic similarity between reports as a source of information, though this information may be slightly less useful than relationships constructed from corporate descriptions. Indeed, reports from different sectors may use similar ESG terminology to describe vastly different corporate contexts.

The Sector-Based Graph shows the lowest performance. While being in the same sector represents an intuitive relationship between companies, this information is less discriminating than the one considered by the other strategies. Constructing the sustainability graph based solely on the industrial sector of companies generates relatively coarse relationships that fail to capture ESG differences between companies in the same industrial sector.

Looking at the graph architecture performances, these results reveal a clear trend. In nearly all configurations, the Graph Transformer is the most effective predictive model. This is due to its ability to use multi-head attention mechanisms and explicitly incorporate edge-associated information during the prediction task. Thus, it can differentiate the importance of connections and leverage company relationships more effectively. GraphSAGE generally ranks as the second-best model. It performs particularly well with the Description-Based Graph, achieving an F1-Score of 0.84, which is essentially equivalent to the Graph Transformer's score. In contrast, the GAT and GCN models perform worse. Although they benefit from the sustainability graph, they appear to have a more limited ability than the Graph Transformer to model complex relationships between companies. GCN consistently performs the worst in most experiments.

Table \ref{tab:results2} shows the results of replacing the embeddings generated by the fine-tuned FinBERT with those produced by the fine-tuned Jina. In all configurations, when using the Jina model, performance decreases compared to FinBERT-generated embeddings. The best Jina-based configuration achieves an Accuracy of 0.81 and an F1-Score of 0.77; these values are lower than those obtained by the best FinBERT-based configuration.

\begin{table}[ht!]
\centering
\scriptsize
\begin{tabular}{@{}llcccc@{}}
\toprule
\textbf{Category} & \textbf{Model / Architecture} & \textbf{Accuracy} & \textbf{Precision} & \textbf{Recall} & \textbf{F1-Score} \\ \midrule

\textbf{Jina fine-tuned embeddings}          & & & & & \\
\textit{Description-Based Graph} & Graph Transformer                 & \textbf{0.79} & \textbf{0.76} & \textbf{0.75} & \textbf{0.76} \\
                           & GraphSAGE                        & \textbf{0.79} & 0.73 & \textbf{0.75} & \textbf 0.74 \\
                           & GAT                              & 0.70 & 0.62 & 0.69 & 0.64 \\
                           & GCN                              & 0.69 & 0.61 & 0.68 & 0.61 \\ \addlinespace

% Gruppo Proposed 2
\textit{Semantic-Based Graph}     & Graph Transformer                 & \textbf{0.81} & \textbf{0.76} & \textbf{0.79} & \textbf{0.77} \\
                           & GraphSAGE                        & 0.78 & 0.73 & 0.75 & 0.74 \\
                           & GAT                              & 0.79 & 0.73 & 0.77 & 0.75 \\
                           & GCN                              & 0.76 & 0.70 & 0.74 & 0.72 \\ \addlinespace

% Gruppo Proposed 3
\textit{Sector-Based Graph}      & Graph Transformer           & \textbf{0.80} & 0.76 & \textbf{0.75} & \textbf{0.75} \\
                           & GraphSAGE                        & 0.78 & 0.72 & \textbf{0.75} & 0.74 \\
                           & GAT                              & 0.79 & \textbf{0.77} & 0.73 & 0.74 \\
                           & GCN                              & 0.75 & 0.70 & 0.71 & 0.70 \\ \bottomrule
\end{tabular}
\caption{Results obtained using Jina-based embeddings instead of FinBERT-based embeddings}
\label{tab:results2}
\end{table}

These results underscore the importance of domain-specific specialization in the context of ESG score evaluation. Although fine-tuned Jina produces high-quality embeddings for general tasks, fine-tuned FinBERT's embeddings incorporate domain-specific knowledge of finance and sustainability, which makes them better suited for ESG classification.

An interesting result to highlight regards the relative performance stability across different configurations. Even when using Jina-based embeddings, the Semantic-Based Graph and the Description-Based Graph continue to outperform the Sector-Based Graph. Meanwhile, the Graph Transformer remains the best-performing predictive model. These results suggest that the superiority of prediction graphs and graph-based predictive models observed in the previous experiments does not depend exclusively on the quality of embeddings but rather represents a more general property of \name{}.

\subsubsection{Single-Dimension Prediction (Environmental, Social, and Governance Scores)}
\label{subsub:Single-Dimension-Prediction}

In addition to evaluating the \name{}'s capability of predicting the overall ESG score, we evaluated the \name{}'s ability to independently estimate the three fundamental dimensions of corporate sustainability, i.e., Environmental, Social, and Governance. To perform this evaluation, we carried out another experiment with two objectives: {\em (i)} verifying the generalizability of \name{} across different sustainability targets, and {\em (ii)} analyzing whether the sustainability graph supporting the graph-based predictive model of \name{} can capture specific information associated with individual ESG components.

To this end, we selected the \name{} configuration that performed best in predicting the overall ESG score. This configuration consisted of a sustainability graph based on company descriptions and the Graph Transformer as predictive model. We used this configuration in the next experiment.

We conducted an independent training process for each ESG dimension. Specifically, we replaced the overall ESG score with the Environmental, Social, and Governance scores provided by the LSEG platform, respectively. We applied the whole methodological pipeline described in Section \ref{sec:Proposed-Approach} for each dimension separately. In particular, we first performed the supervised fine-tuning of FinBERT. Then, we generated the embeddings of sustainability reports and constructed the sustainability graph. Finally, we trained the Graph Transformer.

To assess the impact of sustainability graph, we compared the outcomes of \name{} with those of a fine-tuned FinBERT model that was trained using the same dataset. The latter served as a purely textual baseline. This allowed us to measure the benefit of adding graph learning to purely linguistic representation for each sustainability dimension.

Table \ref{tab:individual_esg_metrics} shows the obtained results. As can be seen, \name{} consistently outperforms the Fine-Tuned FinBERT across all metrics and ESG dimensions. Very good results are obtained for the Environmental component, where Accuracy, Precision, Recall and F1-Score are always greater than 0.85. Furthermore, \name{} improves the Accuracy (resp., Precision, Recall, and F1-Score) of the Fine-Tuned FinBERT by 24.29\% (resp., 17.57\%, 27.54\%, and 23.94\%). These results suggest that information from both sustainability reports and graph-modeled company relationships effectively captures aspects concerning environmental policies, emissions, and resource management.

\begin{table}[]
\centering
\scriptsize
\begin{tabular}{@{}llcccc@{}}
\toprule
\textbf{Dimension} & \textbf{Model / Architecture} & \textbf{Accuracy} & \textbf{Precision} & \textbf{Recall} & \textbf{F1-Score} \\ \midrule

\textit{Environmental} & & & & & \\
                       & \name{} (Desc.-Based Graph+Graph Transf.)                  & \textbf{0.87}     & \textbf{0.87}            & \textbf{0.88}         & \textbf{0.88}           \\
                       & Fine-Tuned FinBERT (baseline)              & 0.70              & 0.74                     & 0.69                  & 0.71                    \\ \midrule

\textit{Social}        & & & & & \\
                       & \name{} (Desc.-Based Graph+Graph Transf.)                  & \textbf{0.85}     & \textbf{0.82}            & \textbf{0.84}         & \textbf{0.83}           \\
                       & Fine-Tuned FinBERT (baseline)              & 0.68              & 0.67                     & 0.61                  & 0.64                    \\ \midrule

\textit{Governance}    & & & & & \\
                       & \name{} (Desc.-Based Graph+Graph Transf.)                  & \textbf{0.79}     & \textbf{0.80}            & \textbf{0.81}         & \textbf{0.80}           \\
                       & Fine-Tuned FinBERT (baseline)              & 0.57              & 0.59                     & 0.53                  & 0.55                    \\ \bottomrule
\end{tabular}
\caption{Comparative results of \name{} and the Fine-Tuned FinBERT for individual Environmental, Social, and Governance dimensions}
\label{tab:individual_esg_metrics}
\end{table}

\name{} also shows significant performance improvement over the Fine-Tuned FinBERT in the Social component. In fact, it improves the baseline of 25.00\% in Accuracy, 22.39\% in Precision, 37.70\% in Recall, and 29.69\% in F1-Score. These results highlight the value of modeling company relationships in identifying common practices and strategies related to human capital management, corporate social responsibility, and stakeholder relationships.

The Governance component is the most difficult to predict. Although \name{} achieves high performance (with an Accuracy of 0.79, a Precision of 0.80, a Recall of 0.81, and an F1-Score of 0.80), the results are lower than those obtained for the Environmental and Social components. One possible explanation is that Governance information tends to be less descriptive and more closely tied to specific corporate, organizational, and decision-making structures. This makes it more difficult to infer all necessary information from textual content and semantic company relationships alone. In fact, this trend can be also observed in the Fine-Tuned FinBERT. Interestingly, the improvements over the baseline are higher for this component than for the other two; in fact, Accuracy increases by 38.60\%, Precision by 35.59\%, Recall by 52.83\% and F1-Score by 45.45\%.

These results demonstrate that \name{} can not only predict the aggregate ESG score, but also  effectively generalize to individual sustainability dimensions. Furthermore, the systematic improvement over purely text-based models confirms the importance of the information about report and company relationships modeled through the sustainability graph. This highlights that information from related companies is a significant factor in assessing the ESG performance of a company.

\subsection{Hyperparameter tuning}
\label{sub:ablation}

After comparing the various \name{} configurations, we conducted some hyperparameter tuning experiments to analyze the impact of certain design choices on the \name{}'s final performance. Specifically, we examined two key aspects of the pipeline: the strategy for generating document-level embeddings and the criteria for constructing the sustainability graph.

The goal of this analysis was to evaluate the \name{}'s robustness across different configurations and to identify settings that balance visualization quality, sustainability graph structure, and ESG prediction accuracy.

The first hyperparameter tuning test examined the effect of the strategy used to generate embeddings for sustainability reports. As described in Section \ref{sub:Step-2}, \name{} supports two types of segmentation, namely: {\em (i)} {\em dynamic segmentation}, which is based on the actual length of the sustainability report, and {\em (ii)} {\em fixed segmentation}, which uses a predetermined number of chunks. The default segmentation in \name{} is the dynamic one because we hypothesized that it better preserves the semantic structure of reports by avoiding an excessive fragmentation in short reports and a loss of information in long ones.

To test this hypothesis, we compared dynamic segmentation with three fixed-segmentation configurations (allowing 5, 10, and 15 chunks), while keeping all the other components of \name{} unchanged. Table \ref{tab:split} shows the results obtained for the three types of graph construction strategies allowed in \name{} (see Section \ref{sub:Step-3}).

\begin{table}[ht!]
\centering
\scriptsize
\begin{tabular}{@{}llccccc@{}}
\toprule
\textbf{Graph Type} & \textbf{Split Method} & \textbf{Accuracy} & \textbf{Precision} & \textbf{Recall} & \textbf{F1-Score} \\ \midrule

\textit{Description-Based Graph} & & & & & & \\
                                 & Fixed (5 splits) & 0.82 & 0.82 & 0.83 & 0.82 \\
                                 & Fixed (10 splits) & 0.83 & 0.81 & 0.84 & 0.82 \\
                                 & Fixed (15 splits) & 0.84 & 0.83 & 0.83 & 0.83 \\
                                 & Dynamic (Used by \name{}) & \textbf{0.85} & \textbf{0.83} & \textbf{0.85} & \textbf{0.84} \\ \midrule

\textit{Semantic-Based Graph}    & & & & & & \\
                                 & Fixed (5 splits) & 0.81 & 0.79 & 0.83 & 0.81 \\
                                 & Fixed (10 splits) & 0.82 & 0.80 & 0.83 & 0.81 \\
                                 & Fixed (15 splits) & 0.82 & 0.82 & 0.84 & 0.82 \\
                                 & Dynamic (Used by \name{}) & \textbf{0.84} & \textbf{0.80} & \textbf{0.85} & \textbf{0.82} \\ \midrule

\textit{Sector-Based Graph}      & & & & & & \\
                                 & Fixed (5 splits) & 0.80 & 0.78 & 0.84 & 0.80 \\
                                 & Fixed (10 splits) & 0.82 & 0.79 & 0.85 & 0.82 \\
                                 & Fixed (15 splits) & 0.83 & 0.81 & 0.84 & 0.82 \\
                                 & Dynamic (Used by \name{}) & \textbf{0.83} & \textbf{0.83} & \textbf{0.84} & \textbf{0.83} \\ \bottomrule
\end{tabular}
\caption{Impact of report segmentation strategies on the performance of \name{}}
\label{tab:split}
\end{table}

This table shows that dynamic segmentation yields the best performance across all configurations. This improvement is particularly evident in the graphs based on company descriptions and semantic similarity of reports. This behavior suggests that adaptive report segmentation generates more informative embeddings that align better with the semantic structure of sustainability reports. Additionally, the observed increase in performance as the number of fixed chunks increases indicates that a more granular document representation is generally beneficial. However, none of the fixed-segmentation configurations matches the results of the dynamic strategy.

The second hyperparameter tuning study examined the sensitivity of \name{} to the thresholds used to construct sustainability graphs. To carry out this task: {\em (i)} we varied the Jaccard similarity threshold for the graphs constructed from company descriptions (see Section \ref{sub:Step-3}); {\em (ii)} we modified the cosine similarity threshold for the graphs constructed from the semantic similarity of reports.

Table \ref{tab:thresholds} reports the results obtained for several values of the two thresholds. The analysis of this table shows that both configurations have an optimal threshold value, beyond which performance begins to degrade. For sustainability graphs based on company descriptions, the best result is obtained with a Jaccard threshold of 0.30. Lower values introduce less informative connections, while higher values excessively reduce the graph's connectivity.

\begin{table}[ht!]
\centering
\scriptsize
\begin{tabular}{@{}llccccc@{}}
\toprule
\textbf{Graph Type} & \textbf{Threshold} & \textbf{Accuracy} & \textbf{Precision} & \textbf{Recall} & \textbf{F1-Score} \\ \midrule

\textit{Description-Based Graph} & $\tau_J$ & & & & & \\
                                 & 0.29 (Jaccard) & 0.83 & 0.82 & 0.83 & 0.82 \\
                                 & 0.30 (Jaccard) & \textbf{0.85} & \textbf{0.83} & \textbf{0.85} & \textbf{0.84} \\
                                 & 0.31 (Jaccard) & 0.84 & 0.80 & 0.85 & 0.82 \\
                                 & 0.33 (Jaccard) & 0.83 & 0.79 & 0.85 & 0.82 \\ \midrule

\textit{Semantic-Based Graph}    & $\tau_{cos}$ & & & & & \\
                                 & 0.75 (Cosine Similarity) & 0.83 & 0.80 & 0.83 & 0.82 \\
                                 & 0.80 (Cosine Similarity) & \textbf{0.84} & \textbf{0.80} & \textbf{0.85} & \textbf{0.82} \\
                                 & 0.85 (Cosine Similarity) & 0.83 & 0.70 & 0.82 & 0.81 \\ \bottomrule
\end{tabular}
\caption{Impact of graph construction thresholds on the performance of \name{}}
\label{tab:thresholds}
\end{table}

A similar behavior is observed in the graph based on the semantic similarity of reports. The optimal threshold for this graph corresponds to a cosine similarity of 0.80. As in the previous case, thresholds that are too permissive increase structural noise, while thresholds that are too restrictive limit the \name{}'s ability to exploit semantic relationships between reports.

\section{Discussion}
\label{sec:Discussion}

The experimental results demonstrate that predicting ESG scores substantially improves when the semantic content of sustainability reports and existing company relationships are jointly considered. The systematic improvement achieved by graph-based configurations, as opposed to purely text-based ones, suggests that a company's ESG performance cannot be fully characterized by treating each report as an isolated information. Rather, similarities in organizational characteristics, business models, and corporate practices of the companies which the reports refer to provide a relational context that contributes to better interpret the report's information. The \name{}'s ability to generalize to individual Environmental, Social, and Governance dimensions further supports this result, as the graph-based component yields significant improvements over the textual baseline in the prediction of the scores of each single dimension.

\name{} may have significant implications for developing automated ESG assessment systems. First, it uses sustainability reports and company metadata. These sources can be collected and processed on a large scale. This offers a more reproducible and adaptable methodology than entirely proprietary assessment pipelines. Second, the ESGPred's ability to predict both the overall ESG rating and its individual components suggests that the support of company relationship information is not limited to a single classification target. 

In any case, we would like to strongly point out that ESGPred and the related frameworks should be viewed as tools that support analysts' judgment, rather than replace it, especially given the complexity and multidimensional nature of ESG criteria.

\name{} has also some limitations that define its current capabilities. First, we used LSEG ratings as the ground truth. Therefore, \name{} learns to replicate a specific rating methodology. However, given the well-known divergence among ESG providers, it would be desirable to ensure that \name{} can also learn to replicate ratings produced according to different criteria. Additionally, the current formulation categorizes scores into four classes and does not directly model the differences between companies within the same class. Finally, \name{} considers static graphs and a limited set of company relationships. However, ESG performance and company relationships evolve over time. These changes may depend on information not represented in the used reports or metadata, such as supply chain relationships, ownership structures, and financial indicators. These limitations suggest enhancements to \name{}, such as the usage of temporal and heterogeneous sustainability graphs, as well as graph explainability methods that can identify which company relationships contribute most to the final assessment.

\section{Conclusion}
\label{sec:Conclusion}

This paper presented \name{}, a modular, graph-based framework that predicts ESG scores from company sustainability reports. Unlike conventional approaches that rely solely on textual analysis, \name{} combines domain-specific language representations with relational graph learning. This enables the prediction process to leverage the semantic content of sustainability reports and the relationships among companies. 

\name{}'s workflow consists of four main steps: \emph{(i)} supervised fine-tuning of FinBERT on ESG-related documents; \emph{(ii)} generation of embeddings from sustainability reports; \emph{(iii)} construction of sustainability graphs representing different kinds of relationship between sustainability reports and/or the companies to which they refer; and \emph{(iv)} prediction through graph-based learning architectures. A key characteristic of \name{} is its flexibility, which allows the embedding model, sustainability graph construction strategy, graph-based predictive model, and target sustainability metric to be modified and evaluated independently. An experimental analysis conducted on a dataset of sustainability reports and ESG ratings demonstrated the effectiveness of integrating textual and relational information. Across all evaluated configurations, graph-based models outperformed purely textual baselines. These results highlight the importance of modeling latent relationships between companies and suggest that descriptive information is a valuable resource for constructing relational structures (such as sustainability graphs) in ESG assessment tasks.

Future research directions include: {\em (i)} investigating dynamic and temporal graph structures that can support the analysis of the evolution of ESG performance over time; {\em (ii)} integrating additional heterogeneous information sources, such as financial indicators, supply-chain relationships, and ownership networks in the construction of sustainability graphs; and {\em (iii)} exploring graph-based structures and predictive models that can handle multiple kinds of entity and relationship simultaneously. 

%\section*{Acknowledgements}
%We acknowledge the support of the project ``MEraviglIA - Metodologie didattiche inclusive ed Intelligenza Artificiale'' (J11I24000700009) under the PR Marche FSE+ 2021/2027 funded by Regione Marche.

%\section*{Declaration of competing interest}
%The authors declare that they have no known competing financial interests or personal relationships that could have appeared to influence the work reported in this paper.

\section*{Data Availability}
The datasets and code used for our study are publicly available at \url{https://anonymous.4open.science/r/ESGPred-25BE/}.
%\url{https://github.com/giulsx/ESGPred}.

{\footnotesize

\bibliographystyle{plain}
\bibliography{bibliografia,temp}

}

\end{document}